\documentclass[10pt,twocolumn,letterpaper]{article} 

\usepackage{avss}
\usepackage{times}
\usepackage{epsfig}
\usepackage{graphicx}
\usepackage{amsmath}
\usepackage{amssymb}

\usepackage{algorithm}
\usepackage{algorithmic}

\usepackage{diagbox}

\usepackage[pagebackref=true,breaklinks=true,letterpaper=true,colorlinks,bookmarks=false]{hyperref}

\avssfinalcopy 

\def\avssPaperID{62} 

\ifavssfinal\fi
\begin{document}

\title{k-Same-Siamese-GAN: k-Same Algorithm with Generative Adversarial Network for Facial Image De-identification with Hyperparameter Tuning and Mixed Precision Training}

\author{Yi-Lun Pan, Min-Jhih Huang, Kuo-Teng Ding, Ja-Ling Wu, Jyh-Shing Jang\\
Department of Computer Science \& Information Engineering, National Taiwan University, Taiwan\\
No. 1, Sec. 4, Roosevelt Rd., Taipei 10617, Taiwan\\
{\tt\small\{d06922016,r07922049,r07922009,wjl,jang\}@csie.ntu.edu.tw}
}

\maketitle

\begin{abstract}
For a data holder, such as a hospital or a government entity, who has a privately held collection of personal data, in which the revealing and/or processing of the personal identifiable data is restricted and prohibited by law. Then, ``how can we ensure the data holder does conceal the identity of each individual in the imagery of personal data while still preserving certain useful aspects of the data after de-identification?" becomes a challenge issue. In this work, we propose an approach towards high-resolution facial image de-identification, called k-Same-Siamese-GAN, which leverages the k-Same-Anonymity mechanism, the Generative Adversarial Network, and the hyperparameter tuning methods. Moreover, to speed up model training and reduce memory consumption, the mixed precision training technique is also applied to make kSS-GAN provide guarantees regarding privacy protection on close-form identities and be trained much more efficiently as well. Finally, to validate its applicability, the proposed work has been applied to actual datasets - RafD and CelebA for performance testing. Besides protecting privacy of high-resolution facial images, the proposed system is also justified for its ability in automating parameter tuning and breaking through the limitation of the number of adjustable parameters.
\end{abstract}

\section{Introduction}
The protection of facial images is getting higher and higher awareness recently, especially when personal images and videos are easily captured by pervasively equipped high resolution visual devices, such as smart phones and surveillance cameras. These devices did simplify the image- and video- footage capturing tasks a lot; nevertheless, attention should be paid to the misuse of the captured imagery data especially when they are stored on a datacenter. De-identification~\cite{gellman2010deidentification} is one of the basic methods aims at protecting the privacy of imagery data and granting their legal usage, at the same time.

In facial image de-identification problems, there are two kinds of dilemma. One is to make the surrogate images as different from the original image as possible so as to ensure the removal of any personal identifiable attribute. The work~\cite{letournel2015face} presented a facial image de-identification method with expression preservation capabilities based on variational adaptive filters, where the filtering process preserves most of the important facial image characteristics (i.e., eyes, gazed regions, lips and their corners).

The other dilemma is to make surrogate images retain as much structural information in the original image as possible so that the image utility remains. Therefore, the existing facial image de-identification procedures based mostly on the k-same algorithms~\cite{gross2008semi,newton2005preserving} and leveraged some facial feature preserving techniques, such as the Active Appearance Models (AAMs) and the Principal Component Analysis (PCA) method~\cite{meng2014retaining}, to explicitly construct morphing-faces for preserving the facial utility attributes like age, race, and gender, as much as possible. Unfortunately, the results are usually too fake to be applied in real world, due to the corresponding poor visual quality and unnatural appearance. Since only traditional de-identification algorithms are involved in the above words, it is our belief that there is still a large room left for performance enhancing with the aid of advanced Deep Neural Network (DNN) related architectures. Moreover, even if DNN-related algorithms and/or architectures are applied, the associated large time spent and huge amount of memory consuming in GPGPUs become the critical obstacles needed to be conquered.

To tackle the above-mentioned obstacles, in this work, a novel GAN-based facial image de-identification system is proposed, which relies on the formal privacy protection ability provided by the above-mentioned k-Same approaches. To release the above-mentioned obstacle-issues in applying DNNs, we not only designed an effective label-scheme function for generating de-identified image set automatically, but also took appropriate hyperparameter tuning into account. In addition, for enhancing the re-identification ability, a recently proposed image recognition NN, called the Siamese Network~\cite{koch2015siamese}, has been modified and integrated into our kSS-GAN. 



Comparing with related works, the proposed facial image de-identification system provides better privacy protection ability, surrogate image quality, and training performance by leveraging the superiority of the mixed precision training (MPT) technique. The main contributions of this work can then be summarized as follows. (1) A novel GAN-based de-identification algorithm for protecting the privacy and reserving the utility of high-resolution facial images at the same time, is proposed. (2) A novel labelling scheme to automatically generate the de-identified image set, is proposed. (3) By combining with the k-Same algorithm, the guarantees of the proposed system against privacy invading of imagery data, is provided. (4) With the aid of MPT, the training process of the proposed work has been speeded up a lot. (5) By Integrating with appropriate automatic Machine Learning (autoML) toolkits, such as Advisor~\cite{golovin2017google}, better hyperparameters turning is achieved.

\section{related works}
\subsection{k-Anonymity and Face De-identification}

The concept of k-Anonymity was addressed by Sweeney~\cite{sweeney2002k} and had been applied to de-identify the entries of a relational database. Sweeney said a table is conformed to the k-Anonymity (after de-identification) if and only if each sequence of values, for each attribute, in the table appears at least k times. Since there are k copies of the same sequence in the data set, the probability of connecting a single sequence to the original sequence is bounded by 1/k. The k-Anonymity model inspired a series of the so-called k-Same de-identification algorithms. The following present brief reviews of some highly relevant articles to this study and analyze their pros and cons.

Gross \etal~\cite{gross2006model} proposed a model-based face de-identification method. The weaknesses of this work include - (1) the group of surrogates cannot be extended, in other words, k-value is limited, (2) the problem of ghosting effects cannot be well-solved, and (3) the most fatal weakness is that the background information of the raw images must be removed so as to align each image with one another. To overcome these challenges, we leverage the k-Same-M positive property (will be detailed later) and integrating AAM into the k-Anonymity parameter space for keeping the visual information and facial expression related attributes as much as possible. \cite{sweeney2002k} addressed a very useful table-based anonymity scheme. It helps k-Anonymity algorithms provide guarantees to protect privacy with limited number of Quasi-identifiers. However, its main drawback lies in dealing only with specific personal data. That is, it focuses only on the text database and cannot be applied to media dataset directly. In this work, we imitate the idea of the former table-based anonymity algorithm and design a novel labeling scheme for generating de-identified image set, automatically. 

Meden \etal~\cite{meden2018k} proposed the k-Same-Net scheme, whose key idea is to combine the k-Anonymity algorithm with GNN architectures. Although the obtained result is the state-of-the-art in this research area, there are some weaknesses still. First, while selecting the cluster centroids (i.e., similar images), traditional PCA algorithm was used; therefore, certain amount of computation is inevitable. Second, it takes quite a long training time when GNN architecture is applied. Another fatal weakness of this approach is that the original images must be down sampled during training and synthesizing, which impairs the quality of surrogate images a lot.

\subsection{StarGAN and Siamese Net}

StarGAN is a multi-domain image-to-image translation framework, within a single dataset or between multiple datasets, and it has shown remarkable results as reported in~\cite{choi2018stargan}. 
Siamese Net is a special type of NN architectures designed for recognizing one-shot images. Instead of being a model assimilating to classify its inputs, it learns to distinguish between two input images. A Siamese Net~\cite{koch2015siamese} consists of two identical NNs, each taking care of one of the input images. The last layer outputs of the two networks are fed to a contrastive loss function, which is used to calculate the similarity of the two images. During the training, the two sister networks share the weights and optimize the contrastive loss function to make similar images as close as possible, and vice versa.
Hence, we use this network as the designed similarity arbitrator to prevent the Generator from producing images that alike to the original one too much.


%
\subsection{Mixed Precision Training (MPT)}

The cost to train or use the model will be grown along with the depth of the network. Therefore, to increase the value of the designed system in real applications, we have to find some ways to reduce the cost, especially in training. The most intuitive way to build a cost-effective model from its original complex form is to simplify the associated model architecture by pruning and/or quantization; nevertheless, in this way, relinquishing in performance is an inevitable side effect~\cite{micikevicius2017mixed}. Literally, MPT means there are more than one floating-point datatypes involved during model training. With the aid of MPT, we can nearly halve the memory requirement and, on recently available GPUs, speed up 2 to 3 times in the training process. Moreover, integrating MPT into a training process introduces no change to any model structure and produces nearly the same performance as that of its original counterpart.

\begin{algorithm}[tb]
\caption{The operations of kSS-GAN}
\label{alg:k-same-GAN}
\textbf{Input}: Input image set $I$, latent variable $Z$, the number of images to synthesize $k$, the label-scheme $L$, and the Model set $M$
\\
\textbf{Output}: De-identified image set $D$ \\
\begin{algorithmic}[1] %
\STATE Aligning input image with the images in $I$, and then randomly generate an initial surrogate image $I^{0}$.
\STATE Computing the distance between $I^0$ and each one of the other images in $I$.
\STATE Extract the latent variable $z^0$ from image $I^0$
\STATE Activating the proposed Auto Labelling function to generate the corresponding label scheme $L$.
\STATE Concatenate the modified StarGAN (will be detailed in Section~\ref{Siameses-GAN}) and Learning the Mode $M$ through training
\FOR{$i \in I$}
    \STATE Loading the pre-trained mode $M$
    \STATE Applying Group ID to Siamese-GAN
    \STATE Generating the de-identified face image $D_i$, and adding it to $D$
\ENDFOR
\end{algorithmic}
\end{algorithm}

\section{System Implementation}
Before discussing the proposed system in detail, as illustrated in Figure~\ref{fig:schematic-fig}, an explanation of the architecture of kSS-GAN for face de-identification, is given. First of all, Radboud Faces Database (RaFD)~\cite{langner2010presentation} and CelebA~\cite{liu2015faceattributes} datasets are chosen as the input image sets for generating new groups of image sets. Due to the concatenation with the modified loss function of Siamese-GAN, the auto label-scheme helps generate a label for each image, according to some specific features, such as facial expression, gender, group id and so on. Finally, we can get the k-same/k-anonymity de-identified faces as the system outputs.
\subsection{The Implementation of kSS-GAN}

\begin{figure*}[tb]
\begin{center}
\includegraphics[width=0.6\linewidth]{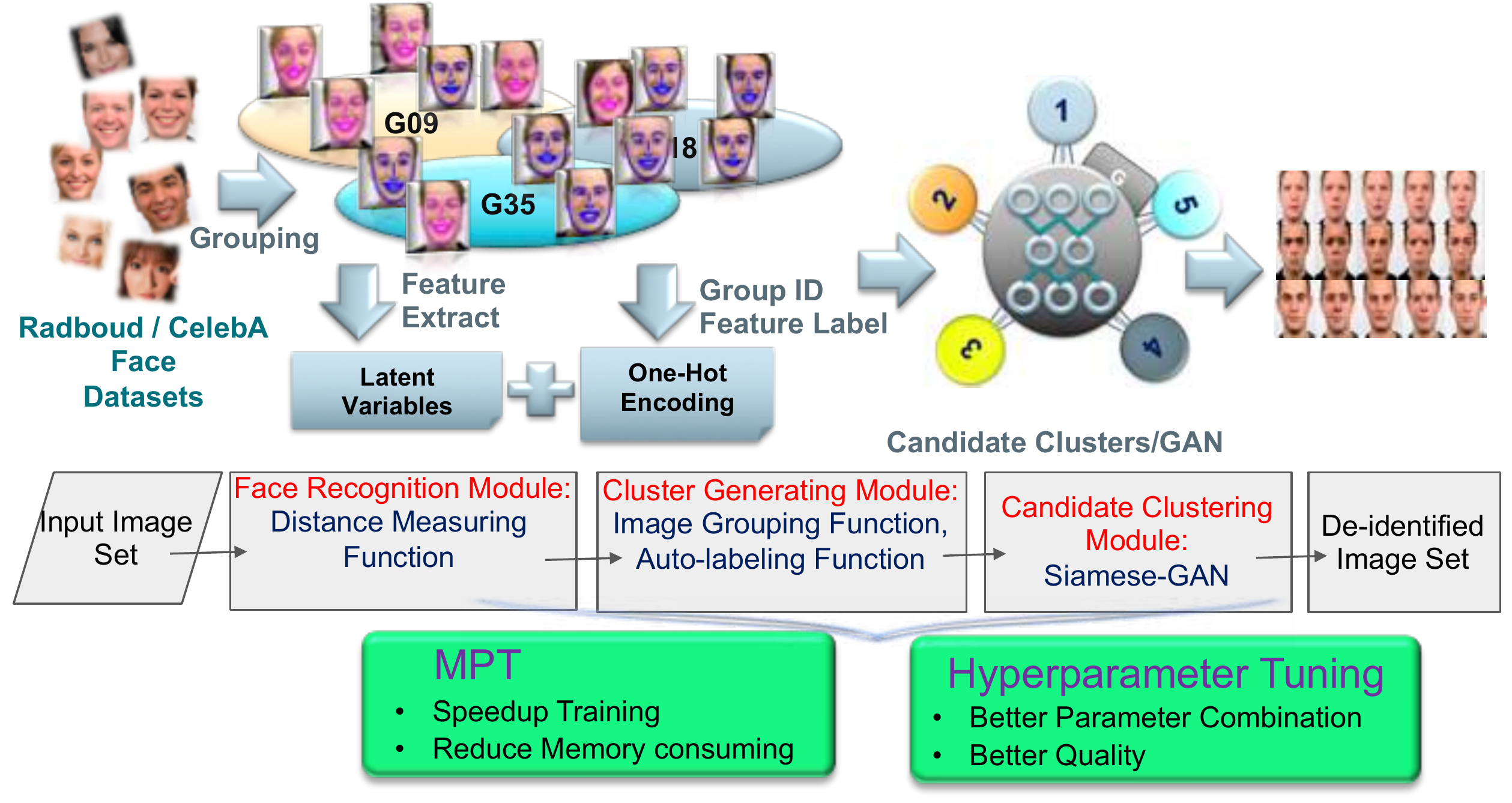}
\end{center}
   \caption{The Schematic Diagram of the proposed kSS-GAN.}
\label{fig:schematic-fig}
\end{figure*}

\begin{algorithm}[tb]
\caption{Siamese-GAN with gradient penalty with gradient penalty. The following default settings: $\lambda_{gp}$ = 10, $\lambda_{rec}$ = 10, $\lambda_{cls}$ = 1, $n_{critic}$ = 5, $\alpha$ = 0.0001, $\beta_{1}$ = 0.5, $\beta_{2}$ = 0.999 are adopted}
\label{alg:SiameseGAN}
\textbf{Input}: The gradient penalty coefficient $\lambda_{gp}$, the domain reconstruction loss coefficient $\lambda_{rec}$, the domain classification loss coefficient $\lambda_{cls}$, the number of critic iterations per generator iteration $n_{critic}$, the batch size $m$, and Adam hyperparameters $\alpha, \beta_1, \beta_2$.\\
\textbf{Parameter}: initial critic parameters $w_0$, initial generator parameters $\theta_0$.
\begin{algorithmic}[1] %
\WHILE{$\theta$ has not converged}
\FOR{$t = 1, ..., n_{criti}$}
    \FOR{$i = 1, ..., m$}
        \STATE Sample real data $x \sim \mathbb{P}_r$, extent latent variable $z \sim p(z)$, generate a random number $\epsilon \sim U [0, 1]$.
        \STATE $\hat{x}\leftarrow \epsilon{x} + (1 - \epsilon)\tilde{x}$
        \STATE{
			$\begin{aligned}
			L_{D}^{(i)} \leftarrow D(\tilde{x}) - D(x) &+ \lambda_{gp}({\|\nabla_{\hat{x}}D(\hat{x})\|_2 - 1})^2\\ &+\lambda_{cls}\mathbb{E}_{x,c'}\big{[}{-\log{D_{cls}(c'|x)}}\big{]}
			\end{aligned}$
        }
    \ENDFOR
    \STATE $w \leftarrow Adam(\nabla_\theta{\frac{1}{m}}\sum_{i=1}^{m}L_{D}^{(i)},w,\alpha,\beta_{1},\beta_{2})$
\ENDFOR
\STATE Sample a batch of latent variables $\big\{{z^{(i)}}\big\}_{i=1}^{m}\sim{p(z)}$, real data $x \sim \mathbb{P}_r$
\STATE $\tilde{x}\leftarrow G_\theta(z)$
\raggedright
\STATE $L_{G}\leftarrow D(x)-D(\tilde{x})+\lambda_{cls}\mathbb{E}_{x,c}\big[{-\log{D_{cls}(c|G(x,c))}}\big]+S(f_{real},f_{fake})$
\STATE $\theta \leftarrow Adam(\nabla_\theta{\frac{1}{m}}\sum_{i=1}^{m}L_{G}^{(i)},w,\alpha,\beta_{1},\beta_{2})$
\ENDWHILE
\end{algorithmic}
\end{algorithm}

kSS-GAN consists of three functional modules: Face Recognition module, Cluster Generating module, and Candidate Clustering modules. The task of Face Recognition module is to align faces and call a distance measuring function to calculate the similarity of the target image to the whole image set. The Cluster Generating module is used to control the image grouping function for selecting k-identity faces and call the auto-labelling function to replace or add specific features for generating candidate clusters. Thus, this module generates the required group ID and labels appropriate scheme (such as facial expressions, color of hairs, and so on) for one-hot encoding. This is the reason why, with the aid of auto-labelling scheme, we can generate de-identified image set automatically and flexibly. The Candidate Clusters module is responsible for communicating with Siamese-GAN to conduct training. At the same time, the hyperparameter tuning tool is activated to find the most appropriate parameter combinations, and MPT is activated as well to speed up the training process and reduce the memory consumption. The schematic diagram of the whole proposed system is sketched in Figure~\ref{fig:schematic-fig}.

The operations of kSS-GAN and Siamese-GAN are detailed in Algorithm~\ref{alg:k-same-GAN} and Algorithm~\ref{alg:SiameseGAN}, respectively. Each image is sampled from the input data and denoted as $x \sim {P_r}$, where $P_r$ is the associated probability. The latent variables $z \sim p(z)$ of each sampled image can be obtained, similarly. The latent variables combined with uniformly distributed random numbers are treated as the input and fed to kSS-GAN. In fact, the target image at the input layer of kSS-GAN is generated based on the latent variables with random numbers $\epsilon \sim U[0,1]$ and the one-hot encoding; therefore, direct controlling the strength of image features at different scales. Combining with noises injected directly into the kSS-GAN network, this kind of architectural arrangement leads to automatic, unsupervised separation of high-level attributes (e.g., pose, identity, facial expressions) from stochastic variations (e.g., freckles, hair) in the generated images, and enables intuitive scale-specific mixing and interpolation operations.

\subsection{The Implementation of Siamese-GAN} \label{Siameses-GAN}

\begin{figure}[ht]
\begin{center}
\includegraphics[width=1\linewidth]{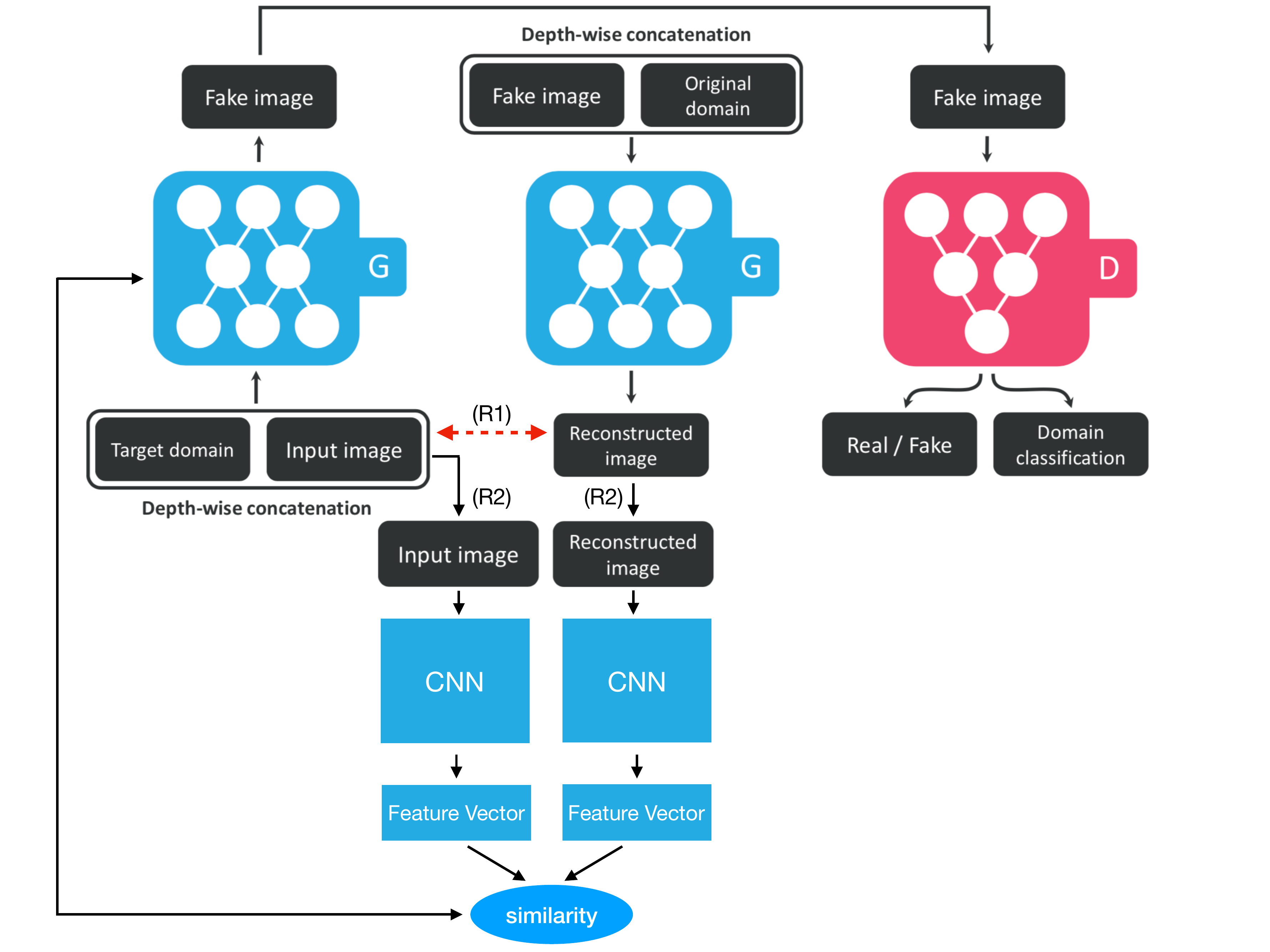}
\end{center}
   \caption{The Information Flow of the Reconstruction Processes.}
\label{fig:siameseGAN}
\end{figure}

In Star-GAN, a generated image must go through the reconstruction process, as shown in Figure~\ref{fig:siameseGAN} (R1), to ensure its relevance to the original one. The authors of Star-GAN then applied Cycle Consistency Loss, defined in \cite{kim2017learning,zhu2017unpaired} to the generator, which can be written as

\begin{equation}
\label{eq:1}
    \mathcal{L}_{rec} = \mathbb{E}_{x,c,c'}\left[ {\left \| x - G(G(x,c),c') \right \|}_1 \right]
\end{equation}
where $G$ takes the translated image $G(x,c)$ and the original domain label $c'$ as inputs and tries to reconstruct the original image x. The $\bold{L}_1$ norm was chosen as the reconstruction loss in Equation~(\ref{eq:1}).

In our work, however, the generated images must be deviated from the original one so as to achieve the so-called K-Anonymity. The newly proposed Siamese-GAN is depicted in Figure~\ref{fig:siameseGAN}, where (R1) stands for the information flow of the reconstruction process of Star-GAN and (R2) denotes that of the newly proposed Siamese-GAN. In (R2), we feed input and reconstructed images into Siamese-GAN to generate the corresponding feature vectors. Then, the calculated similarity is used to guide GAN producing a fake image which is different from the original one. 



In the rest of this section, the involved feature-based similarity measure, reconstruction loss, adversarial loss, domain classification loss, and the overall loss function will be detailed in sequence.

\subsubsection{Feature-Based Similarity}

To overcome the shortages associated with Star-GAN, we modify its reconstruction process from (R1) to (R2), as depicted in Figure~\ref{fig:siameseGAN}, for ensuring that the translated image do distinguishable from the original one. Further, we also apply the Cycle Consistency Loss to the generator and use the feature-based similarity kernel~\cite{wang2017deeplist} to measure the reconstruction loss, that is

\begin{equation}
    \label{eq:2}
    S(f_{real}, f_{fake}) = \frac{1}{2}(1+{f_{real}}^\mathbf{T} f_{fake})
\end{equation}
where $f_{real}$ and $f_{fake}$ are the feature maps of the real and the fake images, respectively. Since $f_{real}$ and $f_{fake}$ have been $\bold{L}_2$-norm normalized, the inner product ${f_{real}}^\mathbf{T}  f_{fake}$ is actually the Cosine similarity between the two originally unnormalized feature vectors. The Cosine similarity is chosen because it is bounded and its gradient with respect to $f_{real}$ and $f_{fake}$, as compared with other bounded similarity functions like the Bhattacharyya coefficient, can be calculated much easily.

\subsubsection{Reconstruction Loss}

To get the feature vector of an image, as shown in Figure~\ref{fig:siameseGAN}, we include the Siamese model~\cite{koch2015siamese} into the reconstruction process. Siamese model will learn facial features of the real and the fake images during the training process and then project the input and the reconstructed images back to the feature space. The so-obtained feature vectors are fed into the feature-based similarity measuring function, as mentioned in the previous sub-section. On the bases of the calculated similarity and the following new reconstruction loss function, cf. Equation~(\ref{eq:3}), a similarity score of the two images will be obtained.

\begin{equation}
    \label{eq:3}
    \mathcal{L}_{rec} = \mathbb{E}_{x,c,c'}\left[ S(x, G(G(x,c), c')\right]
\end{equation}

\subsubsection{Adversarial Loss}

On the other hand, to make the generated (fake) images look close to the original (real) ones, we introduce an adversarial loss function, as given in Equation~(\ref{eq:4}), where $G$ generates an image $G(x, c)$ according to the input image $x$ and the target domain label $c$, while $D$ works to distinguish the real image from the fake one. The whole process behaves like a minmax game. That is, the generator $G$ attempts to minimize the distinguishable objective function, while the discriminator $D$ tries its best to maximize it.

\begin{equation}
    \label{eq:4}
    \mathcal{L}_{adv} = \mathbb{E}_{x}\left[ \log \left( D(x) \right) \right] + \mathbb{E}_{x,c} \left[ \log \right(1 - D(G(x,c))\left) \right]
\end{equation}

\subsubsection{Domain Classification Loss}

For a given input image $x$ and a target domain label $c$, we have to translate $x$ into an output image $y$ which will be correctly classified into the target domain with label $c$. To achieve this goal, we add an ancillary classifier into $D$, just like Star-GAN, and impose two domain-specific classification loss functions, as defined in Equations~(\ref{eq:5}) and (\ref{eq:6}), to optimize $D$ and $G$, respectively. The domain classification loss function for the discriminator $D$ is defined as

\begin{equation}
    \label{eq:5}
    \mathcal{L}_{real\_cls} = \mathbb{E}_{x,c'} \left[- \log \left( D_{cls}(c'|x) \right) \right]
\end{equation}
while the domain classification loss function for the generator $G$ is defined as

\begin{equation}
    \label{eq:6}
    \mathcal{L}_{fake\_cls} = \mathbb{E}_{x,c'} \left[- \log \left( D_{cls}(c'|G(x,c)) \right) \right]
\end{equation}
where $D_{cls}(c'|x)$ denotes the probability distribution over the domain labels computed by $D$. In other words, $G$ tries to minimize this objective function for generating images that can be classified to the target domain with label $c$ and $D$ learns to classify a real image $x$ into its corresponding original domain with label $c'$.

\subsubsection{Overall Loss Function}

Finally, the loss functions for optimizing $G$ and $D$ can now be written, respectively, as
\begin{equation}
    \label{eq:7}
    L_D  = -L_{adv} + \lambda_{cls}*{L_{real\_cls}}
\end{equation}

and
\begin{equation}
    \label{eq:8}
    L_G  = L_{adv} + \lambda_{cls}*{L_{fake\_cls}} + \lambda_{rec}*L_{rec}   
\end{equation}
where $\lambda_{cls}$ and $\lambda_{rec}$ respectively are hyperparameters controlling the relative importance of domain classification loss and the reconstruction loss, as compared with the adversarial loss. Without loss of generality, we use $\lambda_{cls} = 1$ and $\lambda_{rec} = 10$ in all experiments of this work.

\subsection{The Implementation of MPT}

As pre-described, MPT is used to reduce the training time and memory consumption. The kernels of MPT are the floating-point operation with 16-bit precision (FP16) and the Tensor-Core which can accelerate matrix operations and halve the GPU memory consumption. We first cast the input images to FP16 and ingest forward propagation through the model. After that, we convert the model output to FP32 for evaluating the loss and then scale the loss back to FP16 to cover wider representable range. The casting of forward model output to FP32 is a must for preserving small values that are, otherwise, quantized to zero in FP16.

\begin{figure}[tb]
\begin{center}
    \includegraphics[width=.48\textwidth]{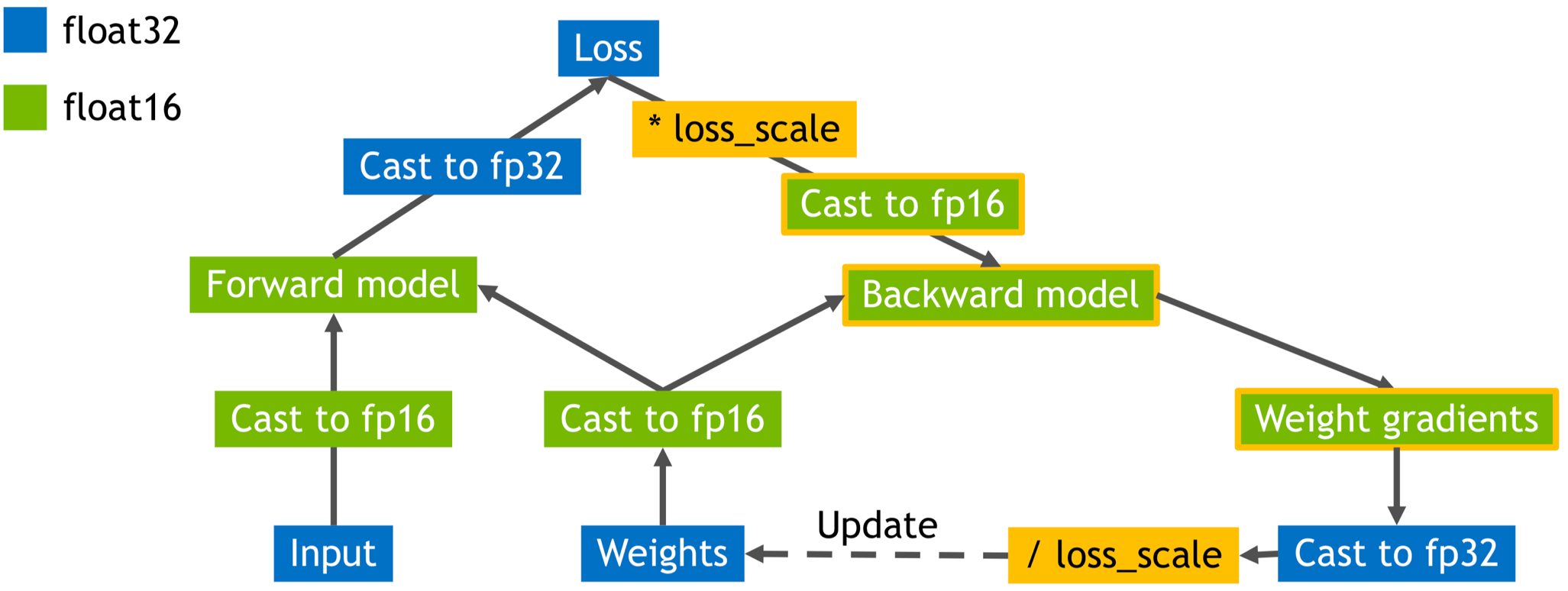}
\end{center}
    \caption{The workflow of the Mixed Precision Tuning.}
    \label{fig:workflow}
\end{figure}

Last but not the least, we have to update the FP32 master-weight and start a new iteration. The FP32 master-weight prevents the updated weight gradients from becoming zero if its magnitude is smaller than $2^{-24}$. The above-mentioned workflow is illustrated in Figure~\ref{fig:workflow}.
\section{The Evaluation of Hyperparameter Performance}

\subsection{Preliminaries of the Experiments and the testing Datasets}


The Radboud Faces Database (RaFD) is used to benchmark the proposed work. RaFD contains high-quality images of 67 subjects, each of which has eight different facial expressions (i.e., anger, disgust, fear, happiness, sadness, surprise, contempt, and neutrality). 
The other tested dataset is CelebFaces Attributes Dataset (CelebA). 
It has large diversities, large quantities, and rich annotations, including 10,177 identities, 202,599 face images, 5 landmark locations, and 40 binary attribute annotations per image. 

\subsection{The Evolution of Hyperparameter Performances}
\begin{figure}[tb]
\begin{center}
    \includegraphics[width=.36\textwidth]{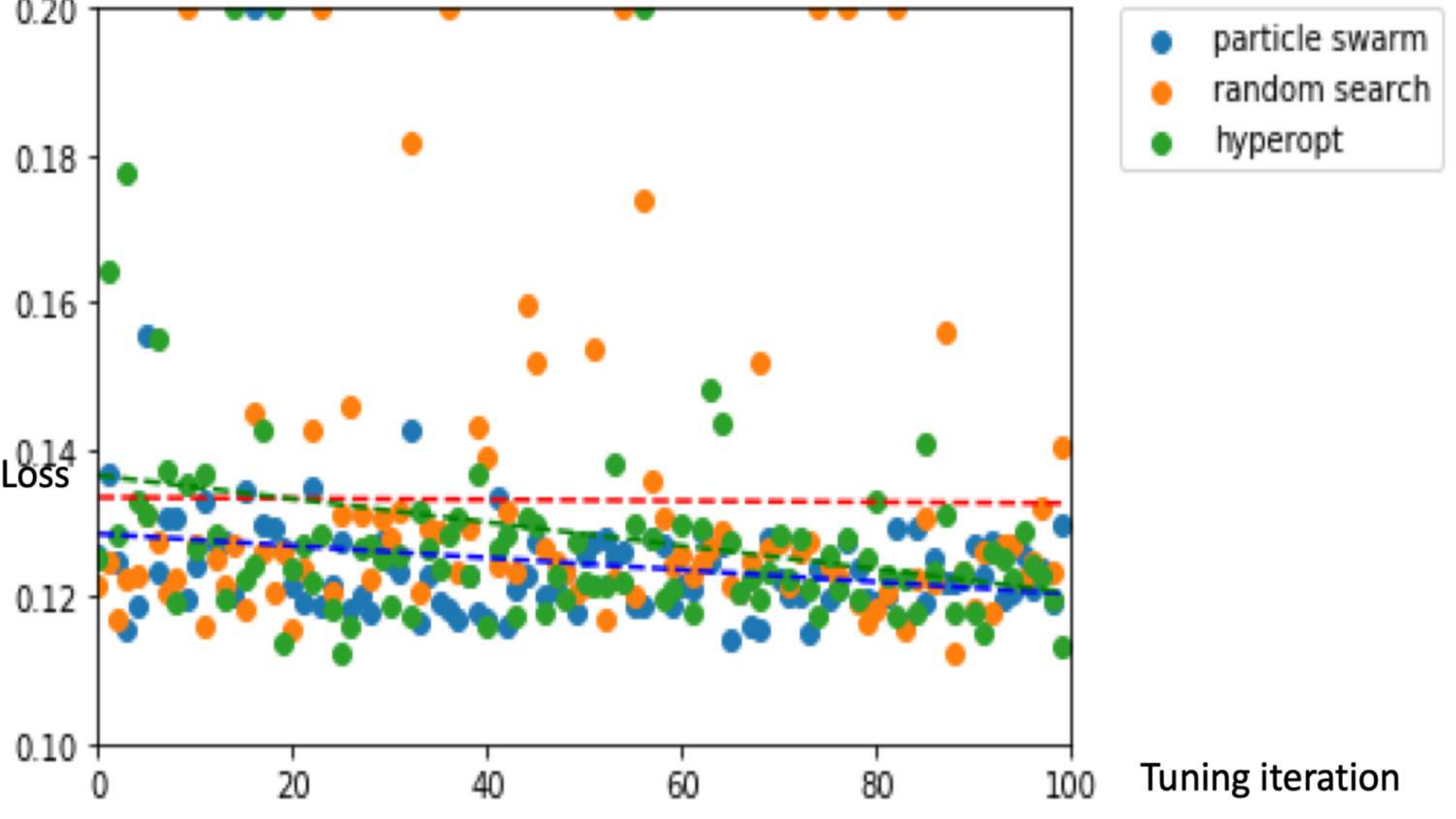}
\end{center}
    \caption{The Corresponding Losses of Hyperparameter Combinations Obtained by Using Three Different Solvers.}
    \label{fig:hyperparameter-losses}
\end{figure}

\begin{figure}[tb]
\begin{center}
    \includegraphics[width=.4\textwidth]{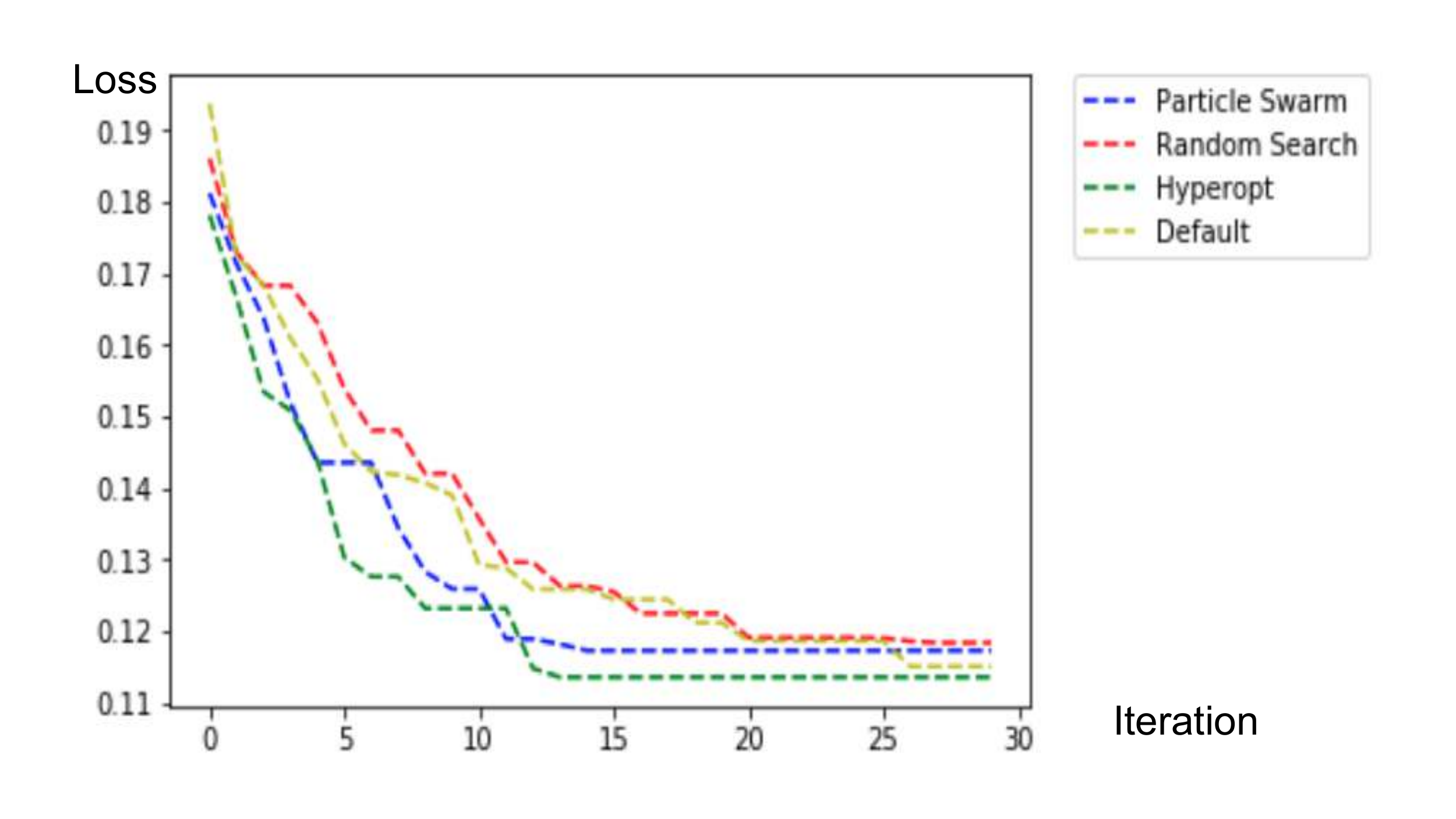}
\end{center}
    \caption{The Overall Losses of the Training Process by Using Four Different Combinations of Hyperparameter.} 
    \label{fig:overall-loesses}
\end{figure}

Three specific solvers of the open-source tool, Advisor, are used to test the proposed system, including Particle Swarm, Random Search, and Hyperopt~\cite{golovin2017google}. Since the Cluster Generator module in kSS-GAN is to choose specific features for generating proper candidate clusters, we use Advisor to search for better hyperparameter combinations. As illustrated in Figure~\ref{fig:hyperparameter-losses}, the x-axe stands for the number of tuning iteration and the y-axe denotes the value of the loss function; clearly, Particle Swarm and Hyperopt provide much lower loss than that of the Random Search. After getting the hyperparameter combination associated with the minimum loss, we use that particular parameter combination to train the model. For comparison, as shown in Figure~\ref{fig:overall-loesses}, we take the loss of the default combination (i.e., without using any solver) into account. Based on the resultant loss of the training process, the combinations of hyperparameters recommended by Advisor do provide lower system loss than that of the native approach.

\subsection{Experimental Results - the effectiveness of MPT}

To illustrate the effects of MPT, we trained the model with the two different GPU platforms listed in Table~\ref{tab:2}, which are Volta V100 and Pascal P100, respectively. The memory consumption before including MPT is quite high; actually, the corresponding memory utility rate is approaching to 100\%. In contrast, by using MPT, the memory utility rates of the system reduce to 36\% for V100 and 37\% for P100, respectively. In MPT, FP16 is one of the computing data types which enabled us to leverage the extremely high speed FP16 matrix operations in Tensor-Core. Roughly speaking, with FP16 computing in Tensor-Core, the training process can be speeded up almost 2 to 3 times, as compared with the case without using MPT in V100. Although not every GPU has Tensor-Core, such as P100, MPT can still reduce lots of memory consumption in training.

\begin{table*}[ht]
\small
\begin{center}
\begin{tabular}{|c|c|c|c|c|c|c|}
\hline
\begin{tabular}[c]{@{}c@{}}\diagbox[]{Device \\ Method}{Performance}\end{tabular} & Speed & \begin{tabular}[c]{@{}c@{}}GPU-RAM\\ Consumption\end{tabular} & Training Time & Speed up & Lightweight\\
\hline\hline
\begin{tabular}[c]{@{}c@{}}Quadro GP100\\ No MPT\end{tabular} & 6 s/10iter & \begin{tabular}[c]{@{}c@{}}9.843GB/16GB \\ (61.5\%) \end{tabular} & 1d 9h & 1 & 0\% \\
\hline
\begin{tabular}[c]{@{}c@{}}Quadro GP100\\ With MPT\end{tabular} & 5-6 s/10iter & \begin{tabular}[c]{@{}c@{}}5.867GB/16GB \\ ($\boldsymbol{36.6\%}$)\end{tabular} & 1d 6h 30m & 1.092 & $\boldsymbol{40.4\%}\downarrow$ \\
\hline
\begin{tabular}[c]{@{}c@{}}Tesla P100\\ No MPT\end{tabular} & 7 s/10iter & \begin{tabular}[c]{@{}c@{}}10.05GB/12GB \\ (83.8\%) \end{tabular} & 1d 14h 54m & 1 & 0\% \\
\hline
\begin{tabular}[c]{@{}c@{}}Tesla P100\\ With MPT\end{tabular} & 7 s/10iter & \begin{tabular}[c]{@{}c@{}}5.971GB/12GB \\ ($\boldsymbol{49.8\%}$)\end{tabular} & 1d 14h 54m & 1 & $\boldsymbol{40.6\%\downarrow}$ \\
\hline
\begin{tabular}[c]{@{}c@{}}GeForce RTX 2080\\ No MPT\end{tabular} & N/A & \begin{tabular}[c]{@{}c@{}} N/A \\ (Run out of Memory) \end{tabular} & N/A & N/A & N/A \\
\hline
\begin{tabular}[c]{@{}c@{}}GeForce RTX 2080\\ With MPT\end{tabular} & $\boldsymbol{4}$ s/10iter & \begin{tabular}[c]{@{}c@{}}6.527GB/8GB \\ (81.5\%)\end{tabular} & $\boldsymbol{22\text{h}}$ & \begin{tabular}[c]{@{}c@{}}GP100 for MPT: $\boldsymbol{1.513\uparrow}$ \\ P100 for MPT: $\boldsymbol{1.768\uparrow}$ \end{tabular} & \begin{tabular}[c]{@{}c@{}}GP100 for MPT: $\boldsymbol{33.6\%\downarrow}$ \\ P100 for MPT: $\boldsymbol{35.2\%\downarrow}$ \end{tabular}\\
\hline
\end{tabular}
\end{center}
\caption{The Characteristics and Performances of the Proposed kSS-GAN.}
\label{tab:2}
\end{table*}

\subsection{The Evolution of kSS-GAN for Face De-identification}
\begin{figure}[tb]
\begin{center}
    \includegraphics[width=.3\textwidth]{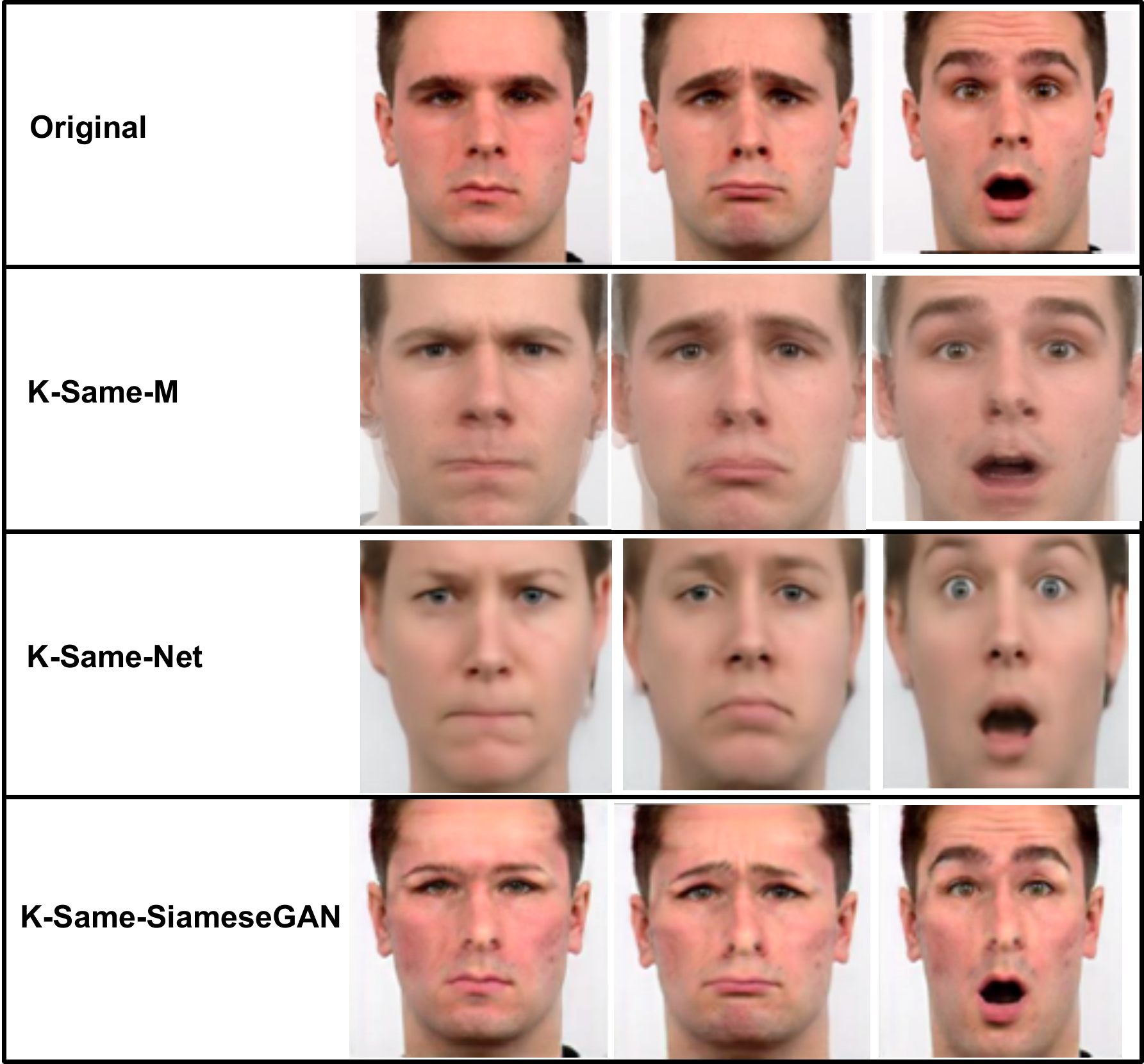}
\end{center}
    \caption{The Evolution of kSS-GAN for Facial Image De-Identification.}
    \label{fig:final-results}
\end{figure}

For checking the effectiveness of kSS-GAN in facial image de-identification, we examine the quality of the fake images generated by K-Same-M, K-Same-Net, and kSS-GAN, respectively. For the same criteria k = 3, we listed the original and the generated images of K-Same-M, K-Same-Net, and kSS-GAN (from top to bottom) in Figure~\ref{fig:final-results}. One should aware of that the schemes differ not only in the generated images’ visual quality but also in the amount of information content retained. In other words, the quality (represented by the naturalness and the smoothness) of the generated image plays the dominating role, if privacy protection (i.e., facial image de-identification) is the only application target. However, under certain application scenarios the identity re-identification is a must, then besides image quality, the amount of retained information associated with the original image becomes very crucial. 

\begin{figure}[!h]
    \centering
    \includegraphics[width=.45\textwidth]{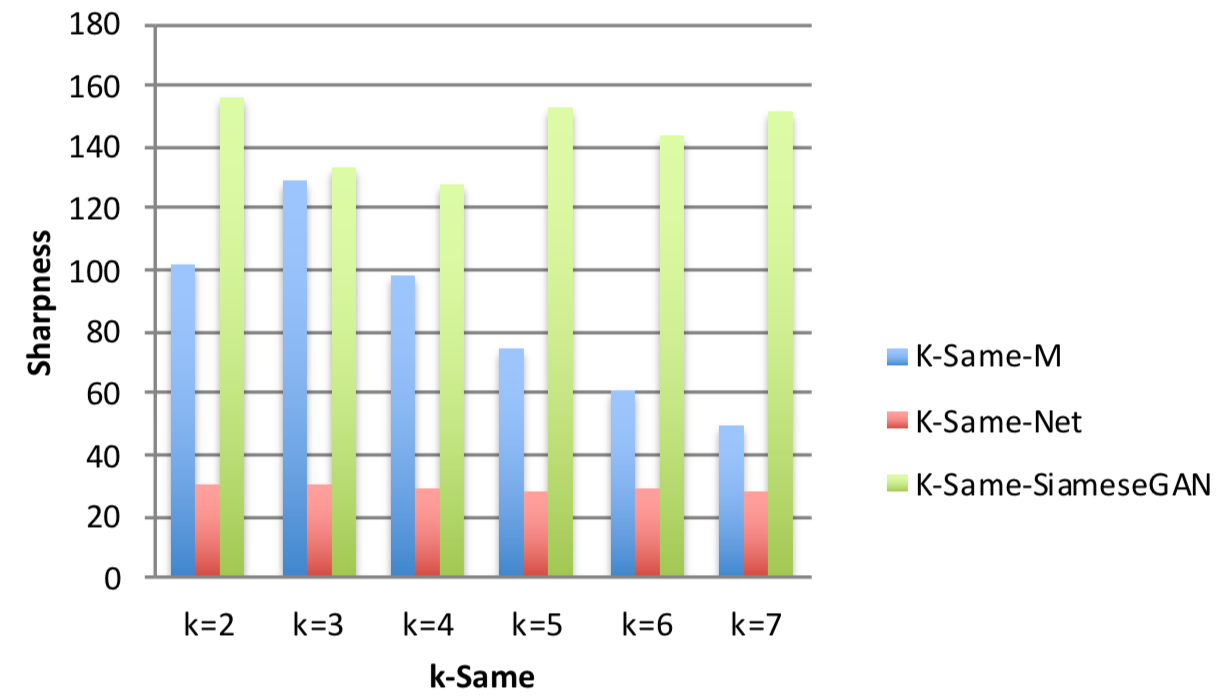}
    \caption{Degrees of Sharpness of the De-Identified Images after Blurry Compression is Applied.}
    \label{fig:blurry-compression}
\end{figure}

In terms of image quality, as shown in Figure~\ref{fig:final-results}, k-same-Net does provide the best result. However, its re-identification capability has never been addressed. The proposed approach, as shown in the bottom of Figure~\ref{fig:final-results}, achieves the goals of providing anonymity guarantee, producing nearly natural and realistic de-identification results, and supporting the ability for re-identifying the original ID if necessary. That is, even though the details of the generated facial image still have some defects, especially in the regions with lights and shadows, which impair the quality of the output image a few. Nevertheless, from Figure~\ref{fig:final-results}, the proposed kSS-GAN provides the highest possibility of ID re-identification from the output images. Moreover, we also find that there are no ghosting effects in the generated images. Thus, we apply Blurry Compression Test to compare the three k-same approaches with the blur detection from OpenCV~\cite{blur2015detection}. Figure~\ref{fig:blurry-compression} shows the comparison results, where the x-axe gives the value of k, and y-axe represents the degree of sharpness (i.e., the variance of Laplacian values). The higher the degree of sharpness is the more the clearness will be. From Figure~\ref{fig:blurry-compression}, the image generated by kSS-GAN is the clearest one in each level of the same k.

\begin{figure}[!h]
\begin{center}
    \includegraphics[width=.45\textwidth]{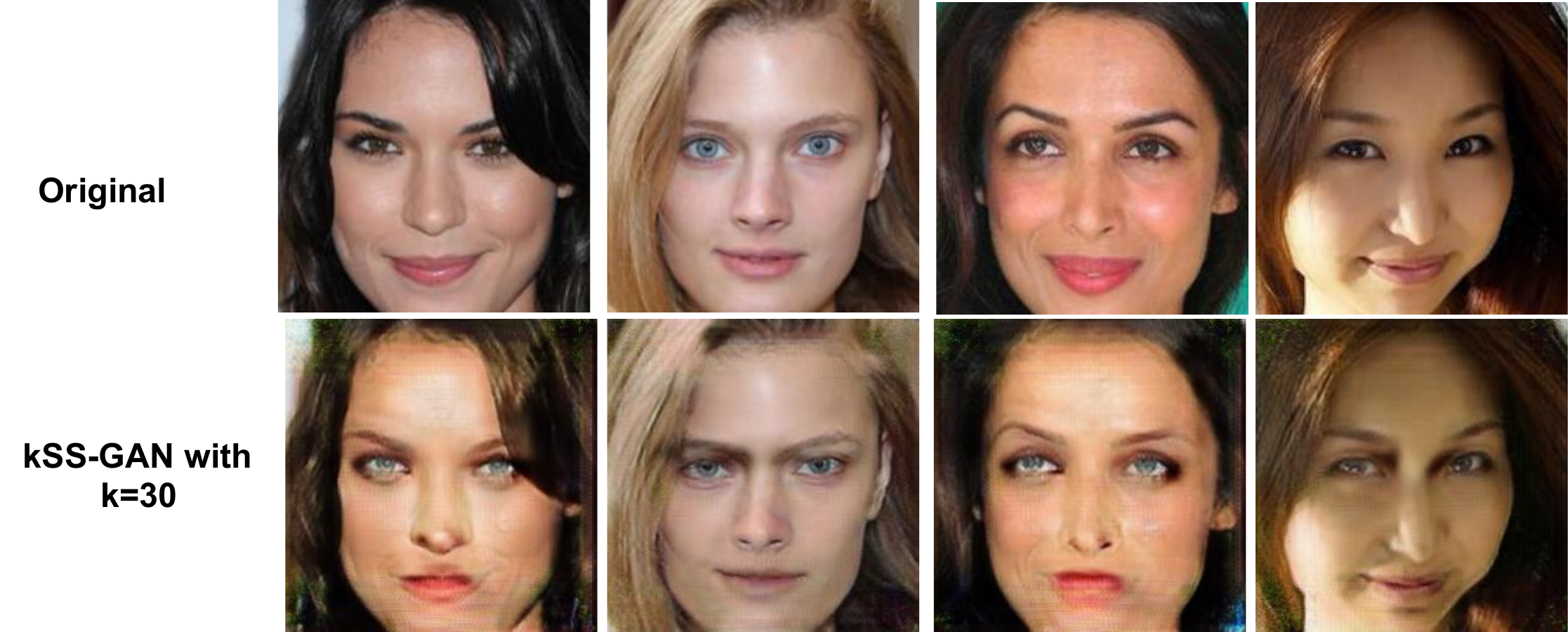}
\end{center}
    \caption{The reconstructed images of kSS-GAN with K = 30.}
    \label{fig:higher-k}
\end{figure}

According to the previous experiments, when the k value is small, the generated surrogate images appear unnatural in the proposed kSS-GAN algorithm. On the other hand, when the value of k is increased to 30, as shown in Figure~\ref{fig:higher-k}, it is found that the naturalness of surrogate images synthesized by the proposed kSS-GAN algorithm has been greatly improved.

\section{Conclusion and Future Work}

\begin{figure}[tb]
    \centering
    \includegraphics[width=.34\textwidth]{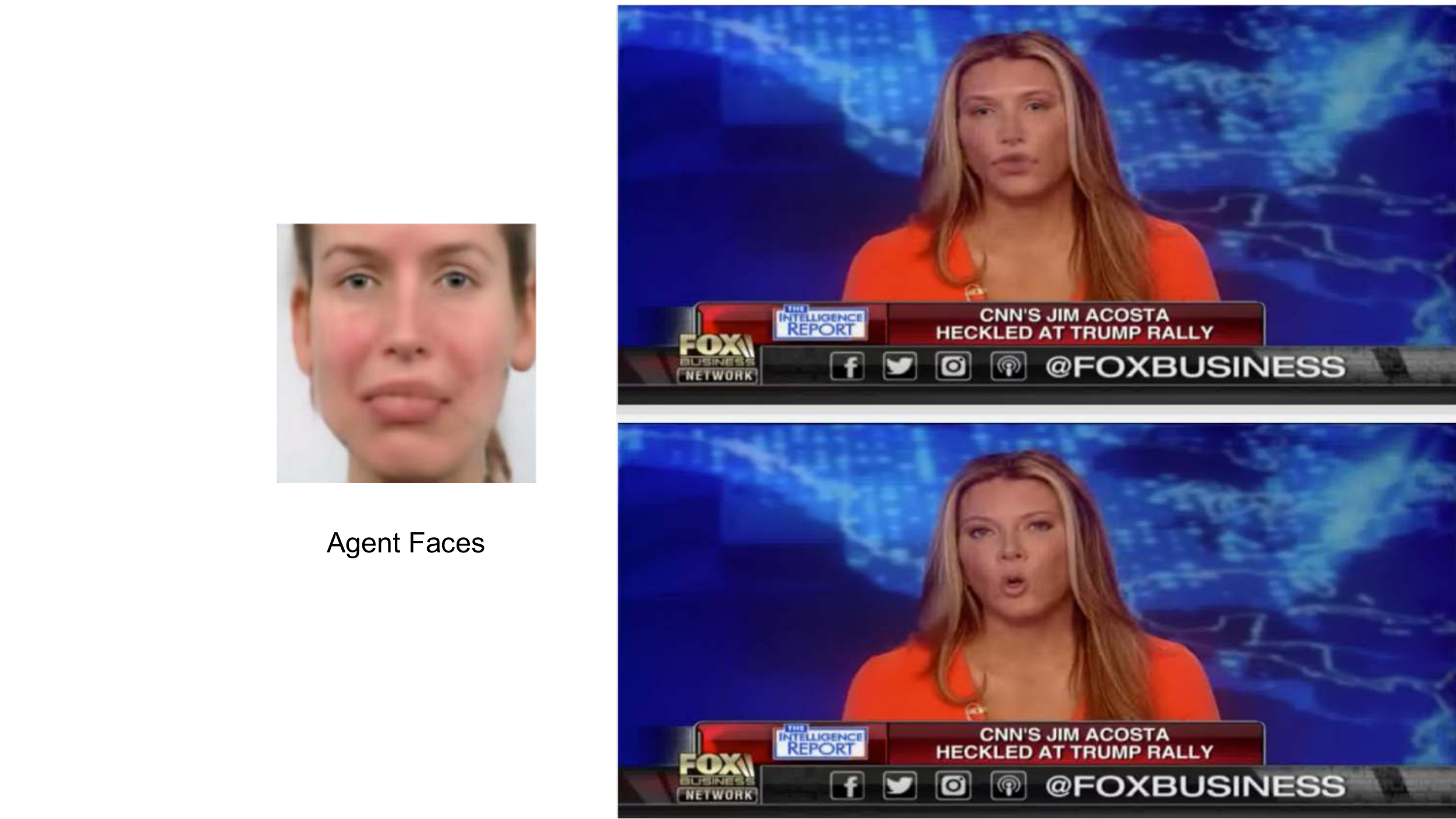}
    \caption{A POC – Prototyping System for Applying Our Work to Video Streams.}
    \label{fig:poc}
\end{figure}

In this work, we proposed a novel approach towards facial image de-identification, called kSS-GAN, which leverages the techniques of k-Same-Anonymity mechanism, GAN and hyperparameter tuning. In order to speed the whole training process up and reduce the memory consumption, the well-performed MPT and the open-sourced hyperparameter tuning tools have also been included in the proposed system. Experimental results showed that kSS-GAN achieved the goals of providing anonymity guarantee, producing nearly natural and realistic de-identification results, and supporting the ability for re-identifying the original ID if necessary. In the future, we will apply the designed approach to video streams for generating the surrogate images in real time; in other words, we will tackle the following critical issues, such as real time object detection from videos, enhancing the quality of images generated by kSS-GAN, and finding effective methods to provide higher model compression. So far, we have realized a proof of concept (POC) prototype system, as shown in the snapshots of Figure~\ref{fig:poc}, when it is applied to a video captured from the youtube (\url{https://youtu.be/MoOpwZme5mc}, \url{https://youtu.be/BgsjZGbJLZ4}).

{\small
\bibliographystyle{ieee}
\bibliography{egbib}

\begin{thebibliography}{10}\itemsep=-1pt

\bibitem{blur2015detection}
{Adrian Rosebrock}.
\newblock Blur detection with opencv.
\newblock
  \url{https://www.pyimagesearch.com/2015/09/07/blur-detection-with-opencv/}.
\newblock Accessed: 2019-05-19.

\bibitem{choi2018stargan}
Y.~Choi, M.~Choi, M.~Kim, J.-W. Ha, S.~Kim, and J.~Choo.
\newblock Stargan: Unified generative adversarial networks for multi-domain
  image-to-image translation.
\newblock In {\em Proceedings of the IEEE Conference on Computer Vision and
  Pattern Recognition}, pages 8789--8797, 2018.

\bibitem{gellman2010deidentification}
R.~Gellman.
\newblock The deidentification dilemma: a legislative and contractual proposal.
\newblock {\em Fordham Intell. Prop. Media \& Ent. LJ}, 21:33, 2010.

\bibitem{golovin2017google}
D.~Golovin, B.~Solnik, S.~Moitra, G.~Kochanski, J.~Karro, and D.~Sculley.
\newblock Google vizier: A service for black-box optimization.
\newblock In {\em Proceedings of the 23rd ACM SIGKDD International Conference
  on Knowledge Discovery and Data Mining}, pages 1487--1495. ACM, 2017.

\bibitem{gross2006model}
R.~Gross, L.~Sweeney, F.~De~la Torre, and S.~Baker.
\newblock Model-based face de-identification.
\newblock In {\em 2006 Conference on Computer Vision and Pattern Recognition
  Workshop (CVPRW'06)}, pages 161--161. IEEE, 2006.

\bibitem{gross2008semi}
R.~Gross, L.~Sweeney, F.~De~La~Torre, and S.~Baker.
\newblock Semi-supervised learning of multi-factor models for face
  de-identification.
\newblock In {\em 2008 IEEE Conference on Computer Vision and Pattern
  Recognition}, pages 1--8. IEEE, 2008.

\bibitem{kim2017learning}
T.~Kim, M.~Cha, H.~Kim, J.~K. Lee, and J.~Kim.
\newblock Learning to discover cross-domain relations with generative
  adversarial networks.
\newblock In {\em Proceedings of the 34th International Conference on Machine
  Learning-Volume 70}, pages 1857--1865. JMLR. org, 2017.

\bibitem{koch2015siamese}
G.~Koch, R.~Zemel, and R.~Salakhutdinov.
\newblock Siamese neural networks for one-shot image recognition.
\newblock In {\em ICML Deep Learning Workshop}, volume~2, 2015.

\bibitem{langner2010presentation}
O.~Langner, R.~Dotsch, G.~Bijlstra, D.~H. Wigboldus, S.~T. Hawk, and
  A.~Van~Knippenberg.
\newblock Presentation and validation of the radboud faces database.
\newblock {\em Cognition and emotion}, 24(8):1377--1388, 2010.

\bibitem{letournel2015face}
G.~Letournel, A.~Bugeau, V.-T. Ta, and J.-P. Domenger.
\newblock Face de-identification with expressions preservation.
\newblock In {\em 2015 IEEE International Conference on Image Processing
  (ICIP)}, pages 4366--4370. IEEE, 2015.

\bibitem{liu2015faceattributes}
Z.~Liu, P.~Luo, X.~Wang, and X.~Tang.
\newblock Deep learning face attributes in the wild.
\newblock In {\em Proceedings of International Conference on Computer Vision
  (ICCV)}, 2015.

\bibitem{meden2018k}
B.~Meden, {\v{Z}}.~Emer{\v{s}}i{\v{c}}, V.~{\v{S}}truc, and P.~Peer.
\newblock k-same-net: k-anonymity with generative deep neural networks for face
  deidentification.
\newblock {\em Entropy}, 20(1):60, 2018.

\bibitem{meng2014retaining}
L.~Meng, Z.~Sun, A.~Ariyaeeinia, and K.~L. Bennett.
\newblock Retaining expressions on de-identified faces.
\newblock In {\em 2014 37th International Convention on Information and
  Communication Technology, Electronics and Microelectronics (MIPRO)}, pages
  1252--1257. IEEE, 2014.

\bibitem{micikevicius2017mixed}
P.~Micikevicius, S.~Narang, J.~Alben, G.~Diamos, E.~Elsen, D.~Garcia,
  B.~Ginsburg, M.~Houston, O.~Kuchaiev, G.~Venkatesh, et~al.
\newblock Mixed precision training.
\newblock {\em arXiv preprint arXiv:1710.03740}, 2017.

\bibitem{newton2005preserving}
E.~M. Newton, L.~Sweeney, and B.~Malin.
\newblock Preserving privacy by de-identifying face images.
\newblock {\em IEEE transactions on Knowledge and Data Engineering},
  17(2):232--243, 2005.

\bibitem{sweeney2002k}
L.~Sweeney.
\newblock k-anonymity: A model for protecting privacy.
\newblock {\em International Journal of Uncertainty, Fuzziness and
  Knowledge-Based Systems}, 10(05):557--570, 2002.

\bibitem{wang2017deeplist}
J.~Wang, Z.~Wang, C.~Gao, N.~Sang, and R.~Huang.
\newblock Deeplist: learning deep features with adaptive listwise constraint
  for person reidentification.
\newblock {\em IEEE Transactions on Circuits and Systems for Video Technology},
  27(3):513--524, 2017.

\bibitem{zhu2017unpaired}
J.-Y. Zhu, T.~Park, P.~Isola, and A.~A. Efros.
\newblock Unpaired image-to-image translation using cycle-consistent
  adversarial networks.
\newblock In {\em Proceedings of the IEEE International Conference on Computer
  Vision}, pages 2223--2232, 2017.

\end{thebibliography}
}

\end{document}